\pdfoutput=1

\documentclass{article}
\usepackage{ijcai25}

\usepackage{times}
\usepackage{soul}
\usepackage{url}
\usepackage[hidelinks]{hyperref}
\usepackage[utf8]{inputenc}
\usepackage[small]{caption}
\usepackage{graphicx}
\usepackage{amsmath}
\usepackage{amsthm}
\usepackage{booktabs}
\usepackage{multirow}
\usepackage{bbding}
\usepackage{subfigure}
\usepackage{graphicx}
\usepackage{amssymb}
\usepackage{algorithm}
\usepackage{algorithmic}
\usepackage[switch]{lineno}
\usepackage{color}

\title{Cause-Effect Driven Optimization for Robust Medical Visual Question Answering with Language Biases}

\author{
Huanjia Zhu$^1$
\and
Yishu Liu$^{2}$\and
Xiaozhao Fang$^{3}$\and
Guangming Lu$^2$\And
Bingzhi Chen$^{1*}$\\
\affiliations
$^1$Beijing Institute of Technology, Zhuhai \\
$^2$Harbin Institute of Technology, Shenzhen \\
$^3$Guangdong University of Technology \\
\emails
bvyih3@gmail.com,
liuyishu@stu.hit.edu.cn,
xzhfang168@126.com,
luguangm@hit.edu.cn,
chenbingzhi@bit.edu.cn$^{*}$\footnote{Corresponding author: Bingzhi Chen.}
}

\begin{document}

\maketitle


\begin{abstract}
    Existing Medical Visual Question Answering (Med-VQA) models often suffer from language biases, where spurious correlations between question types and answer categories are inadvertently established. To address these issues, we propose a novel Cause-Effect Driven Optimization framework called CEDO, that incorporates three well-established mechanisms, i.e., Modality-driven Heterogeneous Optimization (MHO), Gradient-guided Modality Synergy (GMS), and Distribution-adapted Loss Rescaling (DLR), for comprehensively mitigating language biases from both causal and effectual perspectives. Specifically, MHO employs adaptive learning rates for specific modalities to achieve heterogeneous optimization, thus enhancing robust reasoning capabilities. Additionally, GMS leverages the Pareto optimization method to foster synergistic interactions between modalities and enforce gradient orthogonality to eliminate bias updates, thereby mitigating language biases from the effect side, i.e., shortcut bias. Furthermore, DLR is designed to assign adaptive weights to individual losses to ensure balanced learning across all answer categories, effectively alleviating language biases from the cause side, i.e., imbalance biases within datasets. Extensive experiments on multiple traditional and bias-sensitive benchmarks consistently demonstrate the robustness of CEDO over state-of-the-art competitors.
\end{abstract}

\section{Introduction}
Medical visual question answering (Med-VQA) has recently garnered significant attention \cite{liu2021contrastive,liu2022medical}. Med-VQA \cite{SAN2016,MFB2017,BAN2018,chen2022multi,do2021multiple} aims to bridge the gap between medical images and corresponding clinical questions, enabling artificial intelligence systems to predict plausible answers. This automated diagnostic process offers significant advantages, including reduced time expenditure and lower costs, positioning it as a promising technology in the healthcare industry. However, the limited availability of medical datasets \cite{slake2021,vqa-rad2018} poses significant challenges for developing robust Med-VQA. Additionally, inherent biases within Med-VQA have been identified \cite{MICCAI2023}, stemming from the manual splitting and annotation of existing datasets. These biases lead models to learn spurious correlations between question types and answer categories, ignoring critical visual information. Hence, Med-VQA faces a formidable obstacle: \textit{the need to develop robust reasoning capabilities despite the scarcity of medical data, while maintaining performance in out-of-distribution (OOD) training scenarios}.

\begin{figure}[t]
  \centering
  \includegraphics[width=3.35in, height=1.6in]{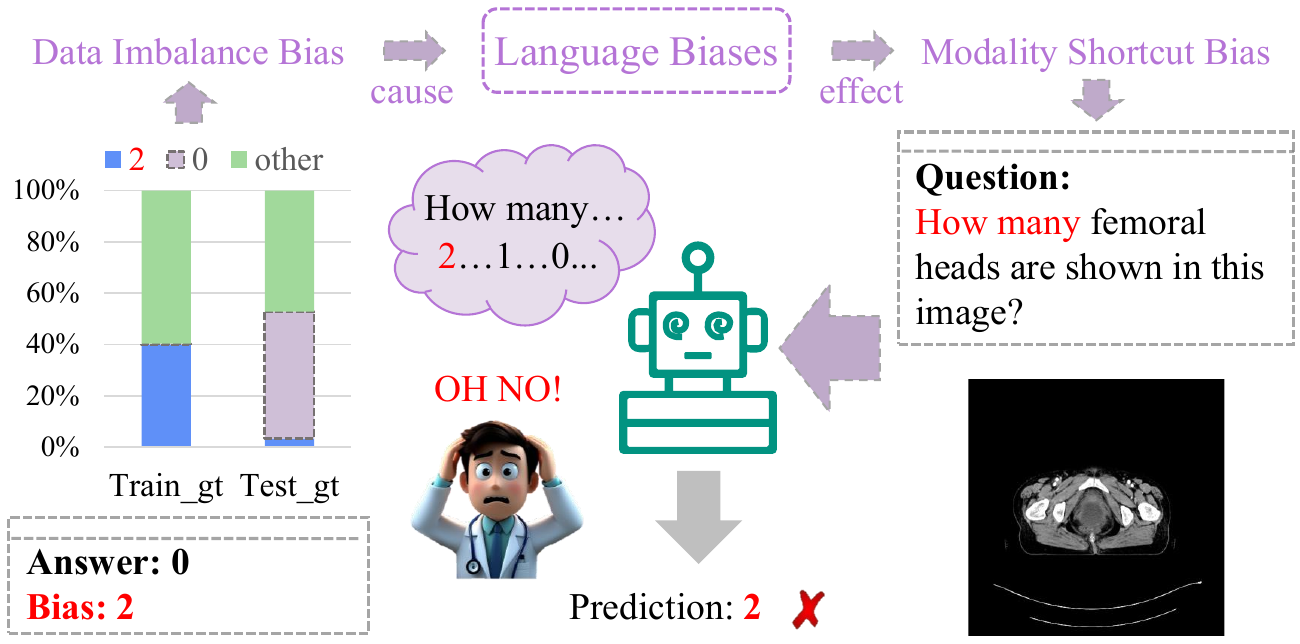}
  \caption{The causal mechanism underlying language biases involves imbalanced label distributions, which contribute to data bias and enable the model to form spurious correlations. For example, the question type ``How many...'' erroneously correlates with specific answers, such as ``2''. This leads to modality shortcut bias, where the model bypasses meaningful reasoning and instead relies on superficial patterns, ultimately generating counterfactual answers.}
  \label{fig1}
\end{figure}

Previous research \cite{MFB2017,SAN2016,BAN2018,chen2022multi,do2021multiple,liu2021contrastive,liu2022medical,ben2019vqa,vu2020question} have adapted general VQA models for Med-VQA tasks. However, these studies often neglect the detrimental impact of language biases, which can lead to critical failures in clinical applications. While recent advances in general VQA \cite{RML2023,GGE2021,GGD2023} have made strides in bias mitigation, such efforts remain underdeveloped in the medical domain. In Med-VQA, counterfactual learning \cite{MICCAI2023} has emerged as a primary approach for bias mitigation, generating counterfactual samples to reduce the model’s dependence on language biases and refocus attention on the target information. However, this approach significantly disrupts the original data distribution, thereby compromising the robustness necessary for Med-VQA.

Addressing language bias requires a comprehensive analysis of its causes and consequences. As illustrated in Figure.~\ref{fig1}, one major cause is \textbf{data imbalance}: frequent answer categories receive disproportionate emphasis during training, leading to overexpansion of their feature space. This amplifies spurious correlations between certain question types and answers.
During training, these spurious correlations are encoded into the network through gradients. Therefore, the model disproportionately relies on the question type for predictions, neglecting critical image features. The question modality gets large gradient updates \cite{reloss2021}, thus becoming a shortcut for generating answers. This phenomenon, driven by language bias, is referred to as \textbf{modality shortcut}. 
By identifying these challenges, this work strives to propose a novel solution to the challenges posed by inherent biases, improving the robustness of Med-VQA models.

To address these challenges, this paper proposes a novel \textbf{C}ause-\textbf{E}ffect \textbf{D}riven \textbf{O}ptimization (\textbf{CEDO}) framework, which aims to alleviate language biases from their cause and effect. Specifically, the proposed CEDO framework innovatively incorporates three mechanisms: \textbf{M}odality-driven \textbf{H}eterogeneous \textbf{O}ptimization (\textbf{MHO}), \textbf{G}radient-guided \textbf{M}odality \textbf{S}ynergy (\textbf{GMS}), and \textbf{D}istribution-adapted \textbf{L}oss \textbf{R}escaling (\textbf{DLR}). 
On the one hand, MHO and GMS are seamlessly integrated to address modality shortcut bias. The primary purpose of MHO is to achieve adaptive optimization for different modalities by adjusting their learning rates. Leveraging the multi-learning rate strategy, it strengthens weaker modalities while preventing dominant modalities from monopolizing the prediction process, thereby mitigating the risk of a single modality becoming a shortcut. Meanwhile, GMS fosters coordinated optimization between modalities through the Pareto method and gradient orthogonality. The Pareto method identifies a steep gradient direction that benefits all objectives (optimizing each modality), allowing the system to converge to a balanced trade-off state. Gradient orthogonality removes gradient conflicts, thereby preventing excessive updates to any single modality and preserving the reasoning capability of other modalities.
On the other hand, DLR mitigates data imbalance bias. It ensures balanced learning across all ground-truth answers by adjusting the loss magnitude for each answer category. Based on statistical data distributions, DLR assigns adaptive weights to rescale individual losses, preventing the model from overemphasizing frequent answer categories at the expense of rare ones.
Our main contributions are summarized as follows:
\begin{itemize}
    \item Our work systematically addresses language biases by targeting their cause and effect. Three innovative mechanisms, i.e., MHO, GMS, and DLR, are proposed to comprehensively address shortcut biases originating from modalities and imbalance biases within datasets.
    \item MHO and GMS cooperate to reduce shortcut biases. MHO enables modality adaptive training, while GMS facilitates coordinated updates between modalities to ensure balanced optimization.
    \item DLR leverages a dynamic loss rescaling strategy to counteract dataset imbalance. By assigning adaptive weights to individual loss, DLR ensures equitable attention across categories.
    \item Two bias-sensitive Med-VQA datasets, SLAKE-CP and VQA-RAD-CP, are built to evaluate the debiasing performance. Extensive experiments on five datasets demonstrate the effectiveness and generalization of our CEDO method, achieving state-of-the-art performance.
\end{itemize}

\section{Related Work}
\subsection{Visual Question Answering}
\paragraph{Medical VQA.}
Existing Med-VQA research \cite{MFB2017,SAN2016,BAN2018,chen2022multi,do2021multiple,liu2021contrastive,liu2022medical,ben2019vqa} typically applies prevalent VQA models to the Med-VQA task, with a primary focus on introducing multimodal feature fusion modules. However, due to the scarcity of medical data, these direct adaptations are often hindered by severe overfitting. To mitigate this challenge, Nguyen et al. \cite{MEVF2019} propose a hybrid enhanced visual feature (MEVF). Despite these efforts, many Med-VQA datasets \cite{slake2021,vqa-rad2018} attempt to balance medical images to alleviate inherent bias dependencies, yet they fail to address the detrimental effects of \textit{OOD} data, thus falling into the trap of language biases.

\paragraph{Robust Medical VQA.}
While research on Med-VQA debiasing has only recently gained traction, studies on bias mitigation in general VQA have flourished. Recent robust VQA approaches focus on introducing bias models \cite{GGE2021,GGD2023} or problem-specific branches \cite{Rubi2019,LMH2019} to learn and mitigate the inherent biases in modalities or datasets. However, due to the scarce medical data, these methods often demonstrate suboptimal generalization performance in the medical domain. As a first attempt, DeBCF \cite{MICCAI2023} specifically targeted this issue and constructed a bias-sensitive dataset to assess debiasing performance. DeBCF employs pre-generated counterfactual samples and counterfactual causal effects to mitigate bias. However, the additional samples cause a distribution shift in the original data, thereby undermining the goal of achieving robust Med-VQA.

\begin{figure*}[t]
  \centering
  \includegraphics[width=0.88\textwidth]{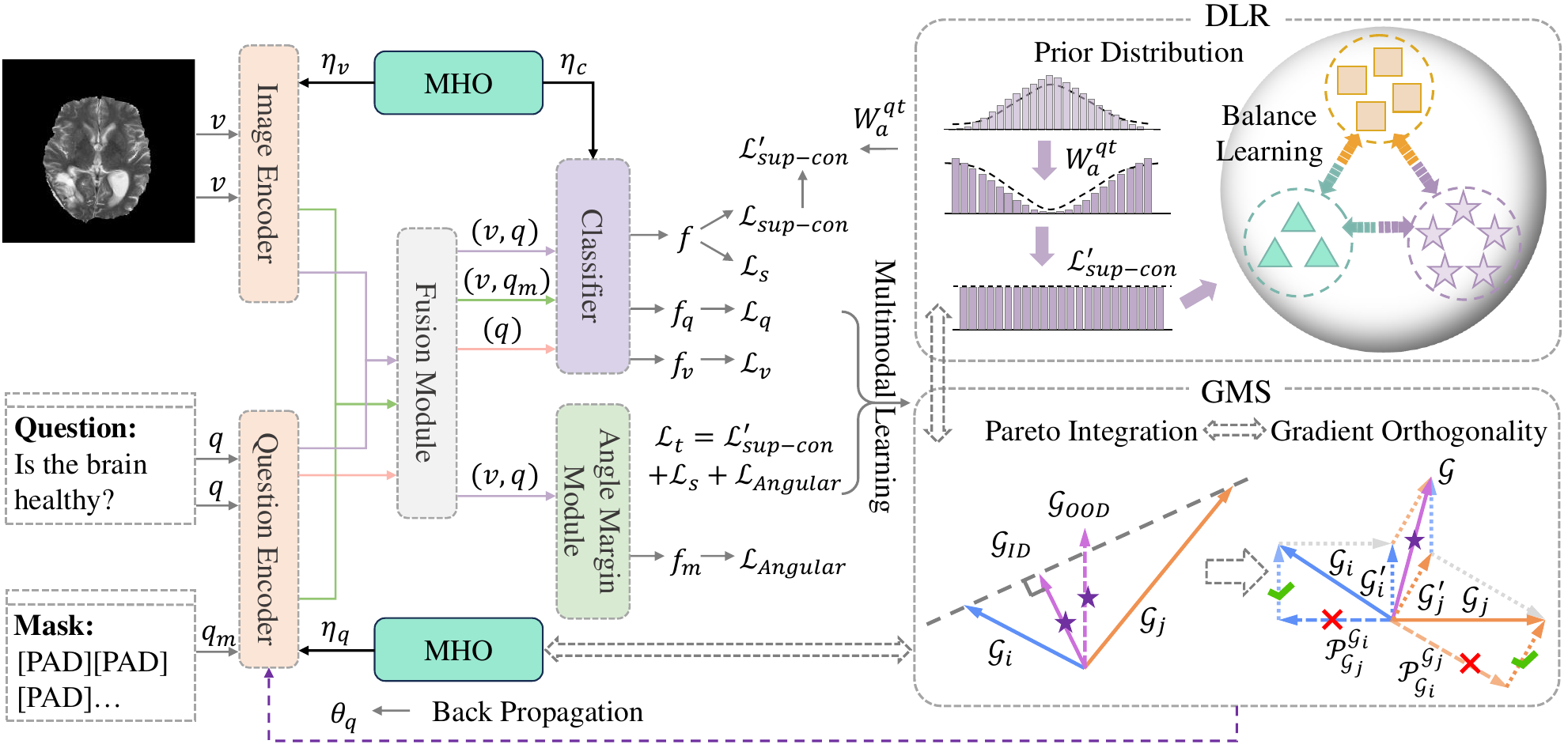}
  \caption{Illustration of the proposed Cause-Effect Driven Optimization (CEDO) framework for addressing medical language biases. Three adaptive mechanisms, i.e., Modality-driven Heterogeneous Optimization (MHO), Gradient-guided Modality Synergy (GMS), and Distribution-adapted Loss Rescaling (DLR), are dexterously established and synergistically integrated to mitigate shortcut bias and imbalance bias from the causal perspective of language biases.}
  \label{fig2}
\end{figure*}

\subsection{Ensemble-Based Methods}
Addressing the elusive and multifaceted nature of bias in VQA tasks, prior research has introduced a question-only model to explicitly capture inherent biases in the dataset and subsequently subtract them from the base VQA model's predictions \cite{Rubi2019,LMH2019,ramakrishnan2018overcoming,GGE2021,GGD2023}. These approaches isolate the question modality to identify and eliminate shortcuts learned by the model.
In addition to these explicit bias removal techniques, data re-weighting methods have emerged as a promising solution. These methods adjust the contribution of individual classes by assigning different weights during loss calculation. A widely used approach is to prioritize tail classes (i.e., less frequent categories) by assigning them higher weights, while head classes (i.e., more frequent categories) receive lower weights, ensuring balanced learning across the dataset.
For example, Focal Loss \cite{ross2017focal} introduces a dynamic weighting mechanism that emphasizes hard-to-classify and misclassified samples, making it particularly effective in imbalanced datasets. Furthermore, Guo et al. \cite{reloss2021} propose a loss re-scaling strategy tailored to the dataset's distribution, further mitigating the overemphasis on dominant classes.

\section{Methodology}

\subsection{Preliminaries}
The Med-VQA task can be viewed as a multi-label classification problem. Without loss of generality, given a batch of data samples $\mathcal{D}$ having $N$ samples, each consisting of an image $v \in \mathcal{V}$, a question $q \in \mathcal{Q}$ and a ground-truth answer $a \in \mathcal{A}$, the goal of the Med-VQA model is to optimize a mapping function $f: \mathcal{V} \times \mathcal{Q} \rightarrow \mathbb{R}^{|\mathcal{A}|}$ to predict the answer corresponding to a given image-question pair. The base Med-VQA model embeds the two features to obtain a joint representation and uses an answer classifier $c$ to generate the logits $f$. Thus the problem can be formulated as follows:
\begin{equation}
    f(v,q) = c( g ( e_{v}(v), e_{q}(q) ) ),
\end{equation}
where $e_{v}$ and $e_{q}$ are the image encoder and question encoder, respectively, and $g$ is a multi-modal feature fusion network. The Med-VQA model can be trained by minimizing the Cross-Entropy (\textit{CE}) loss function:
\begin{equation}
    \mathcal{L}_{CE} = \sum_{i=1}^{|A|} {-a_{i}\log \frac{{\exp(f_{i})}}{\sum_{j=1}^{|A|}{\exp(f_{j})}}}.
\end{equation}
In this paper, we integrate \cite{RML2023} as our base model, and the total loss is denoted as $\mathcal{L}_{t}$.

\subsection{Modality-driven Heterogeneous Optimization}
In Med-VQA, different modalities exhibit distinct complexities and learning requirements. The question modality is prone to biases \cite{reloss2021}, requiring a lower learning rate to avoid overfitting, while the image modality, especially in medical contexts, demands a higher learning rate to capture complex features \cite{khan2022leveraging,liu2021review,tajbakhsh2020embracing,hesamian2019deep}. Besides, the classifier tends to be insufficiently trained \cite{reloss2021}. Traditional single learning rate strategies fail to address these differences, leading to over-reliance on simpler modalities and underutilization of complex ones. To tackle this, we propose MHO, which assigns tailored learning rates to each modality, ensuring synchronized and effective learning aligned with their inherent complexities.

Given a multimodal model $F$, its parameters $\theta$ are grouped into three distinct sets corresponding to the modalities:
\begin{equation}
    \theta = \{ \theta_{q}, \theta_{v}, \theta_{c} \},
\end{equation}
where $\theta_{q}, \theta_{v}, \theta_{c}$ represent the parameters of the question modality, the image modality, and the classifier, respectively. To perform heterogeneous optimization, we define a learning rate $\eta_{k}$ for each modality $k \in \{q,v,c\}$. The total parameter update for a training step is expressed as:
\begin{equation}
\label{eq4}
    \theta_{k} \leftarrow \theta_{k} - \eta_{k} \nabla_{\theta_{k}}L, \quad \forall k \in \{ q,v,c \},
\end{equation}
where $L$ denotes the loss, and $\eta_{q}, \eta_{v}, \eta_{c}$ are the hyperparameters. The MHO mechanism enhances convergence by addressing the inherent differences in modality complexity. By assigning learning rates tailored to each modality, the algorithm prevents overfitting to modality biases (e.g., question shortcuts) and encourages robust learning from underrepresented modalities (e.g., medical images).

\subsection{Gradient-guided Modality Synergy}
A robust Med-VQA model is expected to possess well-trained question and image modalities with aligned optimization directions. However, the pervasive issue of modality shortcut bias has been extensively explored \cite{Rubi2019,GGE2021,GGD2023}, where dominant modalities often exhibit disproportionately larger gradient norms \cite{reloss2021}, leading to imbalanced learning dynamics. Inspired by \cite{pareto,wang2024gradient}, we propose the GMS module that applies the Pareto method and gradient orthogonality to promote synchronized updates between modalities, effectively mitigating the shortcut bias.

\paragraph{Multimodal Learning.}
The Med-VQA task can be viewed as multimodal learning, where models are expected to produce correct predictions by integrating information from multiple modalities. However, only utilizing such joint loss to optimize all modalities together could result in the optimization process being dominated by one modality, leaving others being severely under-optimized \cite{peng2022balanced,huang2022modality}. To overcome this imbalanced multimodal learning problem, introducing unimodal loss, which targets the optimization of each modality, is widely used and verified effective for alleviating this imbalanced multimodal learning problem \cite{wang2020makes}.
Therefore, we introduce a question branch and an image branch using CE loss, denoted as $\mathcal{L}_{q}$ and $\mathcal{L}_{v}$, respectively:
\begin{equation}
    f_{q}(q) = c(g(e_{q}(q) )),
\end{equation}
\begin{equation}
    \mathcal{L}_{q} = \sum_{i=1}^{|A|} {-a_{i}\log \frac{{\exp(f_{q,i})}}{\sum_{j=1}^{|A|}{\exp(f_{q,j})}}},
\end{equation}
where $f_{q}$ denotes the logits in question branch. $f_{v}$ and $\mathcal{L}_{v}$ are computed in the similar way. Thus, the final loss function is:
\begin{equation}
    L = L_{t} + L_{q} + L_{v}
\end{equation}

\paragraph{Pareto Integration.}
In multimodal systems, the relationships between multimodal loss and unimodal losses are intricate, as they are highly interdependent but often exhibit gradient conflicts. These conflicts stem from the fact that optimizing one modality’s loss might detract from the performance of another, especially when modalities vary in representational strength. The gradients of these losses can be expressed as:
\begin{equation}
    \mathcal{G}_{k} = \nabla_{\theta_{q}}\mathcal{L}_{k}(R_{k}, A), \quad \forall k \in \{ t,q,v \},
\end{equation}
where $R_{t}$ represents the joint features, $R_{q}$ and $R_{v}$ correspond to the unimodal features. Resolving how to integrate $\mathcal{G}_{k}$ effectively without introducing additional conflicts is critical. 

The Pareto method, widely utilized in multi-task learning, provides a principled approach for managing these competing gradients \cite{pareto}. It adaptively assigns weights to gradients at each iteration and combines them into a single gradient vector. This ensures that the optimization direction benefits all objectives, facilitating convergence to a Pareto-optimal state. At Pareto-optimality, further improving any single objective is impossible without sacrificing the performance of others.
Integrating Pareto optimization into multimodal learning frameworks naturally aligns with the need to reconcile multimodal and unimodal gradients. The corresponding optimization problem can be formalized as:
\begin{equation}
\begin{gathered}
\min_{\alpha_{t}, \alpha_{q}, \alpha_{v} \in \mathcal{R}}\left\|\alpha_{t} \mathcal{G}_{t}+\alpha_{q} \mathcal{G}_{q} +\alpha_{v} \mathcal{G}_{v} \right\|^{2} \\
\text { s.t. } \quad \alpha_{t}, \alpha_{q}, \alpha_{v} \geq 0, \alpha_{t}+\alpha_{q}+\alpha_{v}=1,
\end{gathered}    
\end{equation}
where $\left\| \cdot \right\|$ denotes the $L_2-$norm. This formulation seeks to minimize the norm of the weighted gradient combination within the convex hull of the gradient family $\{ \mathcal{G}_{k} \}_{k \in \{ t,q,v \} }$. The theoretical properties of this optimization are particularly compelling. As shown in \cite{desideri2012multiple}, there are two key outcomes: (1) if the minimum norm equals zero, the parameters are Pareto-stationary, satisfying a necessary condition for Pareto-optimality; (2) otherwise, the solution identifies a descent direction that is beneficial for all learning objectives. 

\paragraph{Gradient Orthogonality.}
Under an ideal unbiased setting, the Pareto method identifies the optimal gradient direction for each modality, ensuring efficient and balanced updates. However, the inherent bias present in \textit{OOD} datasets significantly influences the model’s learning process, increasing the likelihood of the Pareto method conforming to this bias rather than mitigating it. To address this limitation and suppress shortcut biases, we introduce gradient orthogonality as a corrective mechanism.
Taking $\mathcal{G}_{q}$ and $\mathcal{G}_{v}$ for instance, the cosine similarity $\mathcal{S}$ between two gradients can be written as:
\begin{equation}
    \mathcal{S}_{qv} = \frac{ \mathcal{G}_{q} \cdot \mathcal{G}_{v} } {\Vert \mathcal{G}_{q} \Vert \Vert \mathcal{G}_{v} \Vert}.
\end{equation}
If $S_{qv} > 0$, it indicates that the two modalities are well-aligned during training, with optimization progressing without bias. Conversely, a negative $\mathcal{S}_{qv}$ reveals that $\mathcal{G}_{q}$ and $\mathcal{G}_{v}$ are oriented in opposing optimization directions. In such cases, the modality with a larger gradient norm tends to dominate the training process, leading to shortcut biases. To this end, we employ gradient orthogonality to recalibrate the optimization directions, ensuring unbiased learning dynamics. The projection of $\mathcal{G}_{v}$ onto $\mathcal{G}_{q}$ can be denoted as $\mathcal{P}^{\mathcal{G}_{v}}_{\mathcal{G}_{q}}$:
\begin{equation}
 \mathcal{P}^{\mathcal{G}_{v}}_{\mathcal{G}_{q}} = ( \frac{\mathcal{G}_{v} \cdot \mathcal{G}_{q}} {\Vert \mathcal{G}_{q} \Vert ^{2}} ) \mathcal{G}_{q}. 
\end{equation}
Therefore, the biased gradient norms in the question modality can be dislodged:
\begin{equation}
    \mathcal{G}^{'}_{q} = \mathcal{G}_{q} - \mathcal{P}^{\mathcal{G}_{v}}_{\mathcal{G}_{q}}.
\end{equation}
The gradient norms $\mathcal{G}^{'}_{v}$ and $\mathcal{G}^{'}_{t}$ are obtained in the same manner. In the context of the general case, the ultimate gradient $\mathcal{G}$ can be represented as follows:
\begin{equation}
    \mathcal{G} = \mathcal{G}^{'}_{q} + \mathcal{G}^{'}_{v} + \mathcal{G}^{'}_{t}. 
\end{equation}
This approach ensures effective optimization of the modalities, mitigating shortcut bias arising from the question modality. Notably, the method can also be extended to the image modality or fusion modules, though it necessitates careful consideration of the trade-off between computational resource requirements and potential performance gains.

\subsection{Distribution-adapted Loss Rescaling}
Data imbalance is a primary factor contributing to language biases. Frequent answer categories disproportionately impact the loss function, leading to excessive focus and spurious correlations. To address this issue, inspired by \cite{reloss2021}, we propose an interpretable weighting mechanism that dynamically assigns category-specific weights based on the distribution of question types, ensuring balanced learning across all answer categories.
The weight $w^{j}_{i}$ is obtained via:
\begin{equation}
    w^{j}_{i} = \frac{1}{\mathcal{M}_{j} \times m^{j}_{i}},
\end{equation}
where $\mathcal{M}_{j}$ is the number of samples under question type $qt_{j}$, and $m^{j}_{i}$ is the number of answer ${a_{i}}$ under $qt_{j}$. To achieve a finer balance between rare and common answers, we incorporated the softplus function into the weighting process:
\begin{equation}
    \mathcal{W}^{j}_{i} = \log(1 + \exp{w^{j}_{i}}).
\end{equation}
The softplus function not only smooths extreme weight values but also enhances the stability of the training process. This weighting mechanism is specifically applied to the supervised contrastive loss \cite{khosla2020supervised} in \cite{RML2023}, addressing the adverse effects of imbalanced data on feature spaces in contrastive learning:
\begin{equation}
    \mathcal{L}^{'}_{sup-con} = \sum_{i \in I} \frac{-1}{|P_{i}|} \sum_{p \in P_{i}} \mathcal{W}^{j}_{i} \log \frac{\exp(x^{T}_{i} x_{p}/ \tau)}{\sum_{n \in N_{i}} \exp(x^{T}_{i} x_{n}/ \tau)},
\end{equation}
where $i$ is the index of the current sample in a mini-batch of size $I$ of fused features denoted as $\{ x_{1}, x_{2},...,x_{I} \}$. $a_{i}$ and $qt_{j}$ are the ground truth and question type of the sample $x_{i}$, respectively. The set of positive examples in the mini-batch is represented as $P_{i}:\{ p \in I\ s.t.\ a_{p} = a_{j} \}$, and the set of negative examples is denoted by $N_{i}:\{n \in I\  s.t.\ a_{n} \neq a_{i}\}$. The temperature $\tau$ is set to $1$ following \cite{RML2023}.

The proposed mechanism tackles data imbalance by ensuring balanced contributions across all answer categories, effectively mitigating distributional shifts and improving robustness in \textit{OOD} scenarios. Additionally, this mechanism maintains training stability through moderate weight adjustments while remaining computationally efficient, enabling seamless integration into clinical workflows and widespread adoption without disrupting operational efficiency.

\begin{figure}[t]
  \centering
  \subfigure[SLAKE-CP]{
  \includegraphics[width=3.3in, height=1.6in]{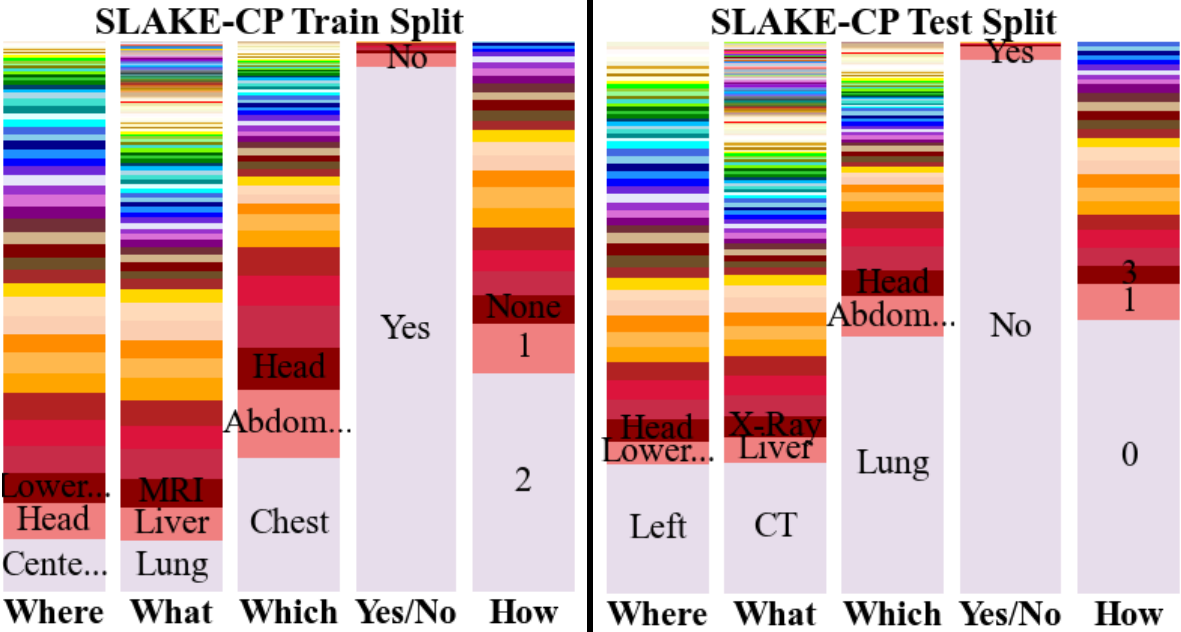}
  \label{Fig3a}}
  \subfigure[VQA-RAD-CP]{
  \includegraphics[width=3.3in, height=1.6in]{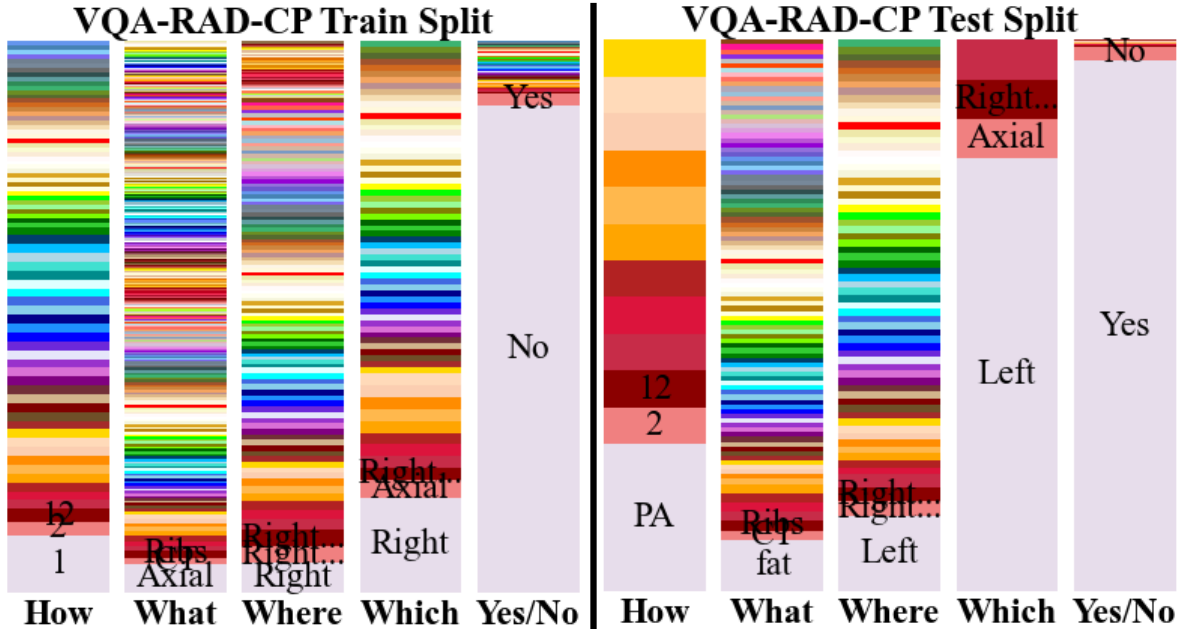} 
  \label{Fig3b}}
  \caption{(a) Designed distribution bias of training set and testing set for all question types in SLAKE-CP. (b) Constructed distribution biases of the training set and the testing set in VQA-RAD-CP.}
  \label{fig3}
\end{figure}

\begin{table*}[t]
  \small
  \centering
  \renewcommand\tabcolsep{8.2pt}
  \renewcommand\arraystretch{0.95}
  \begin{center}
    \begin{tabular}{c|c|c|ccc|ccc}
    \toprule
    \multirow{2}{*}{Approaches} & \multirow{2}{*}{Methods} & \multirow{2}{*}{Reference} & \multicolumn{3}{c|}{SLAKE-CP} & \multicolumn{3}{c}{VQA-RAD-CP	}  \\
    \cline{4-9}
    & & & All & Open & Closed & All & Open & Closed
    \\
    \midrule
     \multirow{4}{*}{Classical}
     & SAN \cite{SAN2016} & CVPR'16 & 26.02 & 48.30 & 6.42 & 16.29 & 59.73 & 6.05 \\
     & MFB \cite{MFB2017} & ICCV'17 & 30.56 & 55.70 & 8.44 & 22.53 & 72.12 & 10.84  \\
     & BAN \cite{BAN2018} & NIPS'18 & 17.50 & 30.90 & 5.72 & 17.30 & 67.70 & 5.42 \\ 
     & UpDn \cite{RML2023} & CVPR'18 & 31.45 & 59.90 & 6.42 & 26.67 & \textbf{74.78} & 15.33 \\
     \hline
     \multirow{2}{*}{Med-Debias}
     & MEVF+SAN \cite{MEVF2019} & MICCAI'19 & 18.62 & 32.60 & 6.33 & 22.11 & 68.14 & 11.26 \\
     & MEVF+BAN \cite{MEVF2019} & MICCAI'19 & 19.33 & 35.00 & 5.54 & 19.07 & 62.39 & 8.86\\ \hline
     \multirow{4}{*}{Natural-Debias}
     & RUBi  \cite{Rubi2019} & NIPS'19 & 33.88 & 60.30 & 10.64 & 81.27 & 60.62 & 86.13  \\
     & LPF \cite{LPF2021} & SIGIR'21 & 40.34 & 43.70 & 37.38 & 41.52 & 65.04 & 35.97 \\
     & GGE-iter \cite{GGE2021} & ICCV'21 & 35.05 & 61.30 & 11.96 & 21.60 & 51.33 & 14.60 \\
     & RMLVQA \cite{RML2023} & CVPR’23 & 76.42 & 60.50 & \textbf{90.41} & 89.45 & 69.03 & 94.26 \\
      \midrule
     \multirow{2}{*}{Ours} & \multirow{2}{*}{CEDO} & —  &\textbf{79.27} & \textbf{66.70} & 90.33 & \textbf{92.07} & 73.45 & \textbf{96.45} \\
     &  &  Increased $\uparrow$ & 2.85 & 6.20 & -0.08 & 2.62 & -1.33 & 2.19 \\
    \bottomrule
    \end{tabular}%
     \caption{Comparisons with state-of-the-art methods are conducted on the SLAKE-CP and VQA-RAD-CP datasets.}
  \label{tab1}%
  \end{center}
\end{table*}%

\begin{table*}[t]
  \centering
  \small
  \renewcommand\tabcolsep{8.2pt}
  \renewcommand\arraystretch{1}
  \begin{center}
    \begin{tabular}{c|c|c|ccc|ccc}
    \toprule
    \multirow{2}{*}{Approaches} & \multirow{2}{*}{Methods} & \multirow{2}{*}{Reference} & \multicolumn{3}{c|}{SLAKE} & \multicolumn{3}{c}{VQA-RAD}  \\
    \cline{4-9}
    & & & All & Open & Closed & All & Open & Closed
    \\
    \midrule
     \multirow{4}{*}{Classical}
     & SAN \cite{SAN2016} & CVPR'16 & 76.00 & 74.00 & 79.10 & 52.89& 31.64 & 65.50 \\
     & MFB \cite{MFB2017} & ICCV'17 & 73.89 & 71.63 & 77.40  &54.10  &41.90 & 62.13  \\
     & BAN  \cite{BAN2018} & NIPS'18 & 76.25 & 75.97  & 76.68 & 55.43& 48.60 & 59.93 \\
     & UpDn \cite{RML2023} & CVPR'18 & 81.34 & 79.84 & 83.65 & 66.74 & 51.40 & \textbf{76.47} \\\hline
     \multirow{2}{*}{Med-Debias}
     & MEVF+SAN  \cite{MEVF2019} & MICCAI'19 & 75.97 & 74.72  & 77.88  & 60.71 & 40.65 &74.05\\
     & MEVF+BAN \cite{MEVF2019} & MICCAI'19 & 77.76 & 75.97 & 80.53  &62.34  &43.09 & 75.14\\ \hline
     \multirow{4}{*}{Natural-Debias}
     & RUBi \cite{Rubi2019} & NIPS'19 & 78.42 & 76.43  & 81.49 & 51.22 & 36.87 & 60.66  \\
     & LPF \cite{LPF2021} & SIGIR'21 & 75.59 & 73.33  & 79.09   & 56.32 &49.72  &60.66 \\
     & GGE-iter \cite{GGE2021} & ICCV'21 & 79.83 & 79.22  & 80.77 & 65.19 & 49.16 & 75.74\\
     & RMLVQA \cite{RML2023} & CVPR’23 & 81.43 & 80.47 & 82.93 & 65.41 & 49.16 & 76.10 \\
      \midrule
    \multirow{2}{*}{Ours} & \multirow{2}{*}{CEDO} & — & \textbf{83.41} & \textbf{81.09} & \textbf{87.02}  & \textbf{67.41} & \textbf{58.66} & 73.16 \\
    &  &  Increased $\uparrow$ & 1.98 & 0.62 & 3.37 & 0.67 & 7.62 & -3.31 \\
    \bottomrule
    \end{tabular}%
      \caption{Comparisons with state-of-the-art methods are conducted on the SLAKE and VQA-RAD datasets.}
  \label{tab2}%
  \end{center}
\end{table*}%

\begin{table}[t]
   \centering
   \small
    \renewcommand\tabcolsep{2.5pt}
    \renewcommand\arraystretch{1.0}
  \begin{center}
    \begin{tabular}{l|c|cccc}
    \toprule
    \multicolumn{2}{l|}{Datasets}  & \multicolumn{3}{c}{VQA-CE} \\
    \midrule
    \multicolumn{2}{l|}{{Methods}} & Overall & Counter & Easy
    \\
    \midrule
    SAN \cite{SAN2016} & CVPR'16 & 55.61 & 26.64 & 24.96 \\
    UpDn \cite{RML2023} & CVPR'18 & 63.52 & 33.91 & 76.69 \\
    RMLVQA \cite{RML2023} & CVPR'23 & 58.05 & 35.01 & 68.21 \\
    MSCD \cite{MSCD2024} & MM'24 & 58.82 & 35.67 & 69.12 \\
\midrule
    CEDO & Ours & \textbf{60.05} & \textbf{35.71} & 71.03 \\
    \bottomrule
    \end{tabular}%
     \caption{The performance comparison on the VQA-CE dataset proves the satisfactory scalability and generalization of the CEDO.}
  \label{tab3}%
  \end{center}
\end{table}%

\begin{table}[t]
  \small
  \centering
  \renewcommand\tabcolsep{6pt}
  \renewcommand\arraystretch{1.0}
    \begin{tabular}{c|ccc|ccc}
    \toprule   
    Methods & MHO & GMS & DLR & All &  Open &  Closed  \\ 
    \midrule
    Baseline & - & - & - & 76.42 & 60.50 & 90.41 \\
    w/ MHO   & \checkmark & - & - & 78.01 & 64.10 & 90.24 \\
    w/ GMS   & - & \checkmark & - & 78.33 & 64.60 & 90.41 \\
    w/ DLR    & - & - & \checkmark & 78.94 & 65.70 & \textbf{90.59} \\
       \midrule
    CEDO & \checkmark & \checkmark & \checkmark & \textbf{79.27} & \textbf{66.70} & 90.33 
\\
    \bottomrule
    \end{tabular}%
  \caption{Ablation experiments for different modules of the CEDO model on the biased SLAKE-CP dataset.}
  \label{tab4}
\end{table}%

\section{Experiments}
\subsection{Datasets}
We evaluate our approach on two classical Med-VQA datasets, SLAKE \cite{slake2021} and VQA-RAD \cite{vqa-rad2018}, along with two \textit{OOD} constructed medical benchmark evaluation protocols, SLAKE-CP and VQA-RAD-CP. To address the scarcity of medical data, we further test our method on a large-scale \textit{OOD} natural benchmark, VQA-CE \cite{vqace}, to verify the scalability and generalization of the proposed approach.
All experiments follow the standard VQA evaluation metric \cite{antol2015vqa}. Implementation details can be found in the supplementary material.

\subsection{Bias Reconstruction}
Following \cite{MICCAI2023}, we propose a novel method to construct two biased Med-VQA datasets, SLAKE-CP and VQA-RAD-CP, designed to serve as valuable benchmarks in future research.
Similar to \cite{MICCAI2023}, the training and test sets were first merged. Each question was labeled based on its question type, determined by its initial words. Samples with binary answers (``Yes'' or ``No'') were categorized as ``Yes/No''. Finally, samples sharing the same question type and answer were grouped into distinct clusters.
As illustrated in Figure.~\ref{fig3}, the samples are redistributed to introduce controlled biases. For each question type, the most frequent answers are allocated to the training and test sets in a 39:1 ratio, while the second most frequent answers are split in a 1:39 ratio. The remaining samples are divided at a 3:1 ratio to ensure the training set contains twice as many samples as the test set, maintaining consistency with the VQA-CP v2 \cite{vqa-cp2018} ratio.

\subsection{Comparisons with State-of-the-art}
\paragraph{Evaluation on Medial Language Biases Benchmark.}
Table~\ref{tab1} demonstrates the superiority of the CEDO framework on the SLAKE-CP and VQA-RAD-CP datasets, specifically designed to evaluate sensitivity to language biases in the medical domain. Key observations include: 1) Most methods exhibit a substantial performance decline on the biased SLAKE-CP and VQA-RAD-CP datasets compared with SLAKE and VQA-RAD. 2) The proposed model outperforms all the state-of-the-art approaches, achieving improvements of 2.85\% and 2.62\% on SLAKE-CP and VQA-RAD-CP, respectively, underscoring its robustness in addressing medical bias challenges. 3) Debiasing models designed for natural scene datasets \cite{Rubi2019,LPF2021,GGE2021} yield suboptimal results, highlighting their limited generalizability to the medical domain.

\paragraph{Evaluation on Medial Standard Benchmark.}
Table~\ref{tab2} highlights the significant performance improvements achieved by our CEDO method on the SLAKE \cite{slake2021} and VQA-RAD \cite{vqa-rad2018} datasets. CEDO demonstrates notable gains of at least 1.98\% and 0.67\% over state-of-the-art approaches, respectively. It is important to note that while many existing methods perform well on in-distribution data, they exhibit pronounced performance degradation when exposed to varying prior conditions.

\paragraph{Evaluation on Natural Standard Benchmark.}
To evaluate the generalization capability of our CEDO approach in realistic application scenarios, we conduct experiments on the challenging \textit{OOD} benchmark, VQA-CE. This benchmark is designed to test the robustness of VQA models under distributional shifts. As shown in Table~\ref{tab3}, the proposed CEDO achieves a minimum improvement of 1.23\%, demonstrating its effectiveness in handling \textit{OOD} conditions.

\begin{figure}[t]
  \centering
  \subfigure[$\eta_{q} \times 1000$]{
  \includegraphics[width=1.1in, height=0.9in]{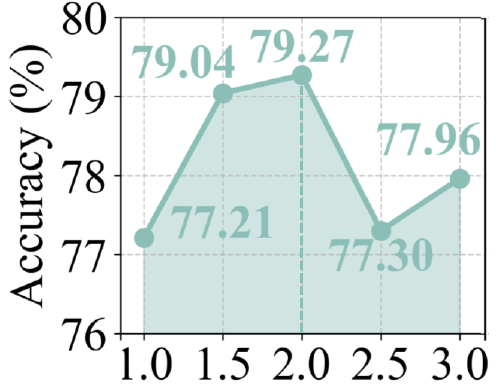}
  \label{Fig4a}
  \hspace{-0.08in}}
  \subfigure[$\eta_{v} \times 1000$]{
  \includegraphics[width=1.1in, height=0.9in]{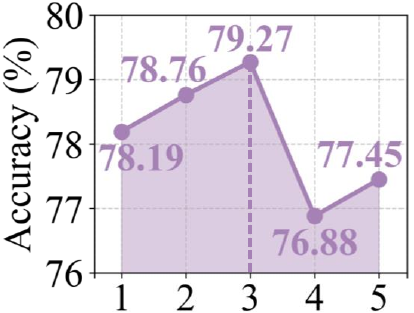}
  \label{Fig4b}
  \hspace{-0.08in}}
  \subfigure[$\eta_{c} \times 1000$]{
  \includegraphics[width=1.1in, height=0.9in]{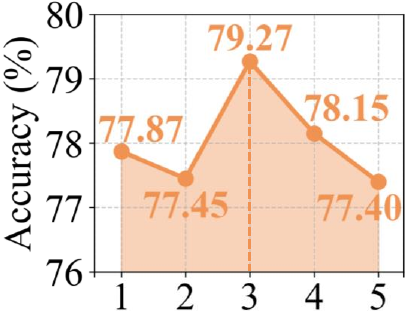}
  \label{Fig4c}
  \hspace{-0.08in}}
  \vspace{-0.1in}
  \caption{Comparison of Accuracy on the SLAKE-CP dataset with different parameter configurations.}
  \label{fig4}
\end{figure}

\begin{figure}[t]
  \centering
  \subfigure[Prior distribution]{
  \includegraphics[width=3.3in, height=1.4in]{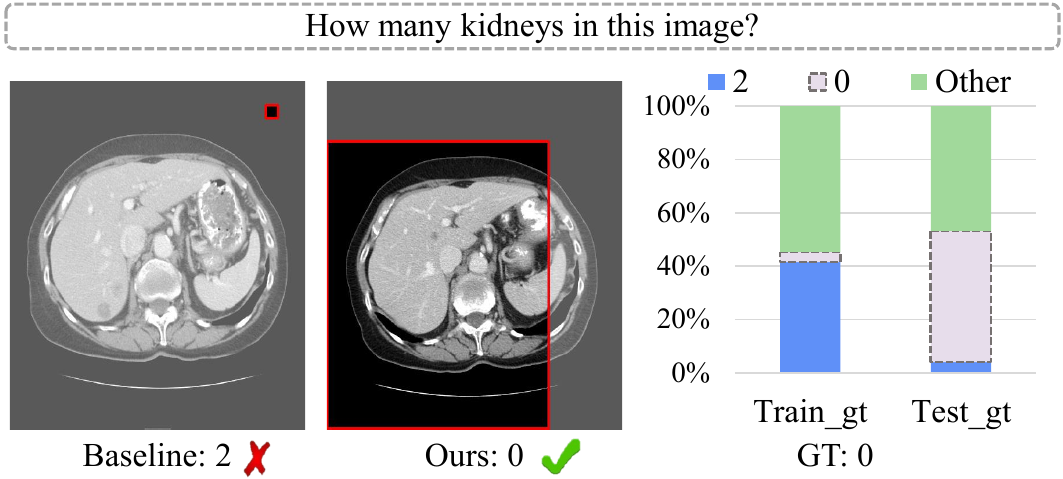}
  \label{Fig5a}
  }
  \subfigure[Rare diseases distribution]{
  \includegraphics[width=3.3in, height=1.4in]{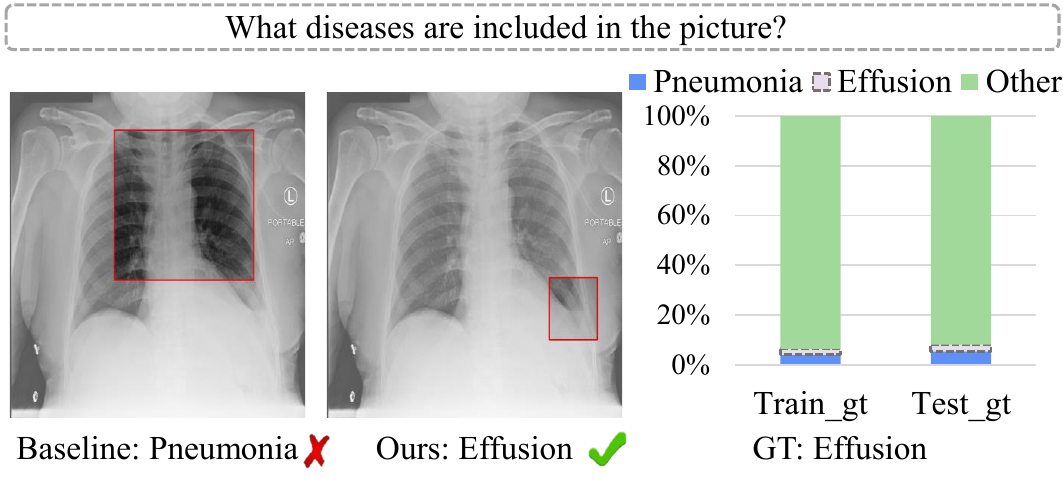}
  \label{Fig5b}
   }
  \caption{Visualization analysis of CEDO. Our proposed approach combines robust reasoning capabilities for rare diseases with an effective strategy for mitigating bias-related challenges.}
  \label{fig5}
\end{figure}

\begin{figure}[t]
  \centering
  \subfigure[How]{
  \includegraphics[width=1.685in, height=1.48in]{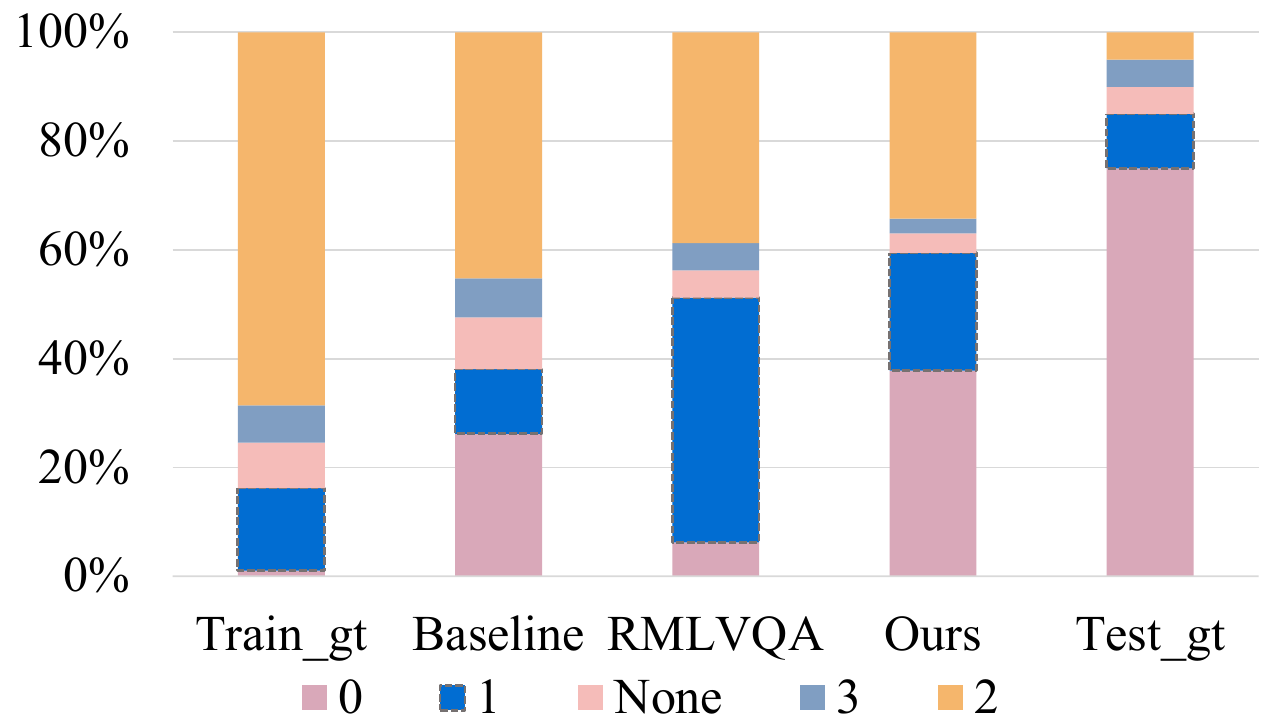}
  \label{Fig6a}
  }
  \hspace{-0.2in}
  \subfigure[What]{
  \includegraphics[width=1.685in, height=1.48in]{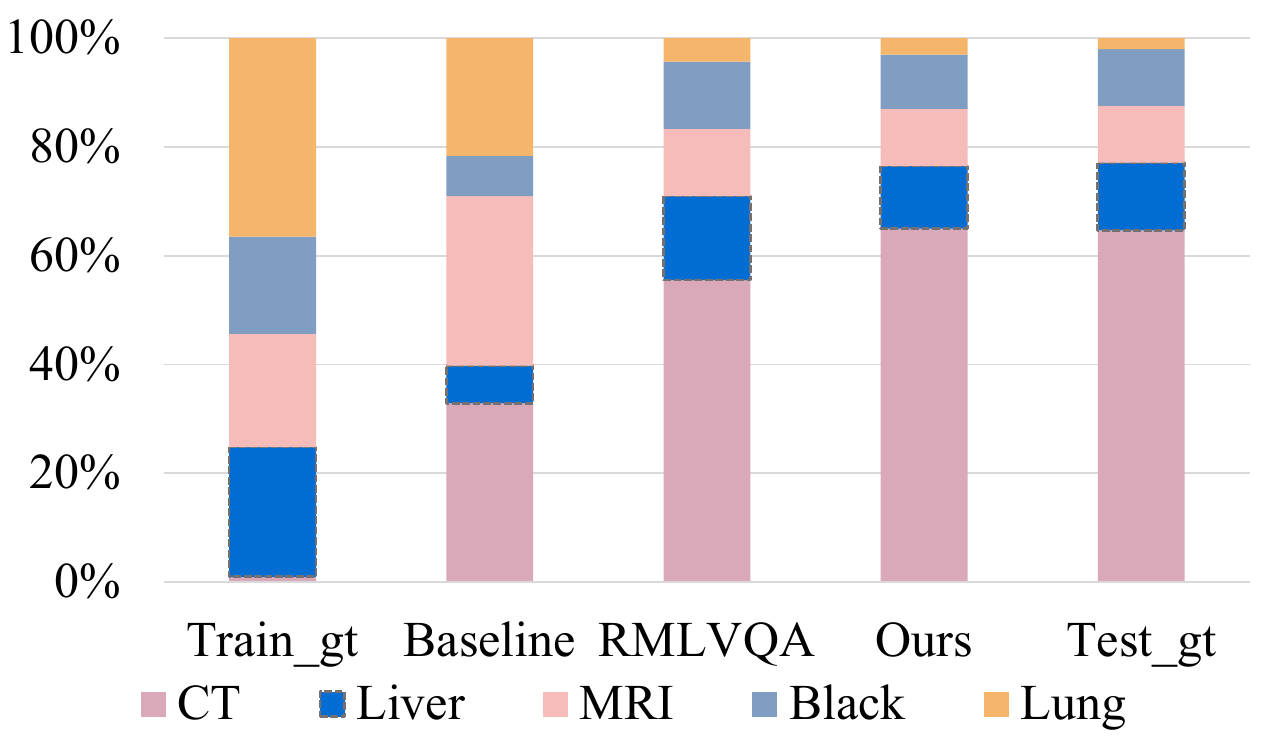}
  \label{Fig6b}
  }
  \caption{Bias Mitigation: Answer distributions of ``How'' and ``What'' question types on SLAKE-CP demonstrate that CEDO maintains prominent improvements over baseline and RMLVQA.}
  \label{fig6}
\end{figure}

\subsection{Ablation Study}
To assess the effectiveness of each component, ablation studies were conducted on SLAKE-CP, as presented in Table~\ref{tab4}. The results are summarized as follows: 1) \textbf{Baseline:} We integrate \cite{RML2023} as the base model. 2) \textbf{Baseline w/ MHO:} Adding the MHO mechanism improves performance by 1.59\%, demonstrating its effectiveness in tailoring learning rates for different modalities. This mechanism mitigates language bias while enhancing the learning of complex visual features, leading to better multimodal integration. 3) \textbf{Baseline w/ GMS:} The GMS component yields a performance gain of 1.91\%, effectively facilitating collaborative optimization between modalities, thus significantly reducing shortcut bias. 4) \textbf{Baseline w/ DLR}: The addition of DLR achieves a notable performance improvement of 2.52\%, highlighting its capability to prevent sparse categories from being neglected and to mitigate data imbalance bias. 5) \textbf{Baseline w/ CEDO:} The full CEDO model achieves the best performance, demonstrating its robustness in tackling language bias comprehensively from both cause and effect sides.

\subsection{Parameter Analysis}
We conduct an in-depth parameter analysis of the proposed CEDO method by exploring its behavior under various hyperparameter configurations. Our investigation focuses on three key hyperparameters: $\eta_{q}$, $\eta_{v}$, and $\eta_{c}$, as specified in Eqn. (\ref{eq4}). Through systematic experimentation and detailed analysis, as depicted in Figure. \ref{fig4}, we observe that the model achieves peak performance when $\eta_{q}=0.002$, $\eta_{v}=0.003$, and $\eta_{c}=0.003$. This analysis highlights that an optimal combination of these hyperparameters can achieve superior performance of the CEDO model.

\subsection{Visualization Results}
From the analysis in Figure~\ref{fig5}, our CEDO method demonstrates dual advantages in Med-VQA by prioritizing critical visual information. It achieves outstanding debiasing performance, reducing language biases and modality shortcuts, while ensuring high diagnostic accuracy for rare diseases—an often overlooked yet vital aspect of medical AI. Additionally, Figure~\ref{fig6} demonstrates that our CEDO method significantly alleviates medical language bias by shifting the focus from biased patterns to essential information within the data. This enables the model to accurately predict unbiased answers, even in challenging scenarios where language shortcuts or distributional biases may otherwise dominate.

\section{Conclusion}
In this paper, we identified the cause and effect of medical language biases, i.e., shortcut bias and imbalance bias. To overcome these challenges, we proposed an innovative Cause-Effect Driven Optimization (CEDO) framework that addresses language biases from a causal perspective. The proposed method introduces a multi-learning rate strategy to heterogeneously optimize each modality, then applies the Pareto method and gradient orthogonal technology to achieve inter-modality coordination, thereby mitigating shortcut bias, combined with the loss rescaling mechanism to alleviate the imbalance bias. Two bias-sensitive Med-VQA datasets are constructed to evaluate the debiasing performance.

\section*{Ethical Statement}

There are no ethical issues.

\section*{Acknowledgments}
This work was supported in part by the Shenzhen Fundamental Research Fund (No. JCYJ20240813105900002), in part by the Guangdong Basic and Applied Basic Research Foundation (No. 2025A1515010225), and in part by the National Natural Science Foundation of China (No. 62302172).

\section*{Contribution Statement}
 $^*$Corresponding author: Bingzhi Chen.

\bibliographystyle{named}
\bibliography{ijcai25}

\end{document}